\documentclass[twoside,11pt]{article}

\usepackage{jmlr2e}
\usepackage{listings}
\usepackage{color}
 
\definecolor{codegreen}{rgb}{0,0.6,0}
\definecolor{codegray}{rgb}{0.5,0.5,0.5}
\definecolor{codepurple}{rgb}{0.58,0,0.82}
\definecolor{backcolour}{rgb}{0.95,0.95,0.92}
 
\lstdefinestyle{mystyle}{
    backgroundcolor=\color{backcolour},   
    commentstyle=\color{codegreen},
    keywordstyle=\color{magenta},
    numberstyle=\tiny\color{codegray},
    stringstyle=\color{codepurple},
    basicstyle=\footnotesize,
    breakatwhitespace=false,         
    breaklines=true,                 
    captionpos=b,                    
    keepspaces=true,                 
    numbers=left,                    
    numbersep=5pt,                  
    showspaces=false,                
    showstringspaces=false,
    showtabs=false,                  
    tabsize=2
}
 
\jmlrheading{7}{2016}{1-5}{7/16}{-}{Guillaume Lema\^itre, Fernando Nogueira, 
and Christos K. Aridas}

\ShortHeadings{A Python Toolbox to Tackle the Curse of Imbalanced Datasets in Machine Learning}{Lema\^itre, Nogueira, and Aridas}
\firstpageno{1}

\begin{document}

\title{Imbalanced-learn: A Python Toolbox to Tackle the Curse of Imbalanced Datasets in Machine Learning}
\author{Guillaume Lema\^itre \email g.lemaitre58@gmail.com \\ 
    \addr{LE2I UMR6306, CNRS, Arts et M\'etiers, Univ. Bourgogne Franche-Comt\'e} \\ 
    \addr{12 rue de la Fonderie, 71200 Le Creusot, France} \\ 
    \addr{ViCOROB, Universitat de Girona} \\ 
    \addr{Campus Montilivi, Edifici P4, 17071 Girona, Spain}
        \AND
        Fernando Nogueira \email fmfnogueira@gmail.com \\ 
        \addr{ShoppeAI} \\ 
        \addr{488 Wellington Street West, Suite 304, Toronto, Ontario M5V 1E3, Canada}
        \AND
        Christos K. Aridas \email char@upatras.gr \\ 
        \addr{University of Patras} \\ 
        \addr{University Campus, 26504 Patras, Greece}} 
\editor{-}

\maketitle

\begin{abstract}%
\texttt{imbalanced-learn} is an open-source python toolbox aiming at providing a wide range of methods to cope with the problem of imbalanced dataset frequently encountered in machine learning and pattern recognition.
The implemented state-of-the-art methods can be categorized into 4 groups: (i) under-sampling, (ii) over-sampling, (iii) combination of over- and under-sampling, and (iv) ensemble learning methods.
The proposed toolbox only depends on \texttt{numpy}, \texttt{scipy}, and \texttt{scikit-learn} and is distributed under MIT license.
Furthermore, it is fully compatible with \texttt{scikit-learn} and is part of the \texttt{scikit-learn-contrib} supported project.
Documentation, unit tests as well as integration tests are provided to ease usage and contribution.
The toolbox is publicly available in GitHub \\ \url{https://github.com/scikit-learn-contrib/imbalanced-learn}.
\end{abstract}


\section{Introduction}

Real world datasets commonly show the particularity to have a number of samples of a given class under-represented compared to other classes.
This imbalance gives rise to the ``class imbalance'' problem~\citep{prati2009data} (or ``curse of imbalanced datasets'')
which is the problem of learning a concept from the class that has a small number of samples.

The class imbalance problem has been encountered in multiple areas such as 
telecommunication managements, bioinformatics, fraud detection, and medical diagnosis,
and has been considered one of the top 10 problems in data mining and 
pattern recognition~\citep{yang200610,rastgoo2016tackling}. 
Imbalanced data substantially compromises the learning process, since most of the 
standard machine learning algorithms expect balanced class distribution or an 
equal misclassification cost~\citep{he2009learning}. For this reason, several
approaches have been specifically proposed to handle such datasets.
Such standalone methods have been implemented mainly in R language~\citep{torgo2010data,kuhn2015caret,dal2013racing}.
Up to our knowledge, however, there is no python toolbox allowing such processing while cutting edge machine learning toolboxes are available~\citep{pedregosa2011scikit,sonnenburg2010shogun}.

In this paper, we present the \texttt{imbalanced-learn} API, 
\textit{a python toolbox to tackle the curse of imbalanced datasets
in machine learning}. The following sections present the project vision, a snapshot of the API, an overview of the implemented methods,
and finally, the conclusion of this paper, including future functionalities
for the \texttt{imbalanced-learn} API.

\section{Project management}

\noindent\textit{Quality insurance} In order to ensure code quality, a set of unit tests is provided leading to a coverage of 99 \% for the release 0.1.8 of the toolbox.
Furthermore, the code consistency is ensured by following \texttt{PEP8} standards and each new contribution is automatically checked through landscape, which provides metrics related to code quality.

\noindent\textit{Continuous integration} To allow user and developer to either use or contribute to this toolbox, Travis CI is used to easily integrate new code and ensure back-compatibility.

\noindent\textit{Community-based development} All the development is performed in a collaborative manner.
Tools such as git, GitHub, and gitter are used to ease collaborative programming, issue tracking, code integration, and idea discussions.

\noindent\textit{Documentation} A consistent API documentation is provided using \texttt{sphinx} and \texttt{numpydoc}.
An additional installation guide and examples are also provided and centralized on GitHub\footnote{\url{https://github.com/scikit-learn-contrib/imbalanced-learn}}.

\noindent\textit{Project relevance} At the edition time, the repository is visited no less than $2,000$ per week, attracting about $300$ unique visitors per week.
Additionally, the toolbox is supported by \texttt{scikit-learn} through the \texttt{scikit-learn-contrib} projects.

\section{Implementation design}

\begin{lstlisting}[language=Python, caption=Code snippet to over-sample a dataset using SMOTE.]
from sklearn.datasets import make_classification
from sklearn.decomposition import PCA
from imblearn.over_sampling import SMOTE

# Generate the dataset
X, y = make_classification(n_classes=2, weights=[0.1, 0.9],
                           n_features=20, n_samples=5000)

# Apply the SMOTE over-sampling
sm = SMOTE(ratio='auto', kind='regular')
X_resampled, y_resampled = sm.fit_sample(X, y)
\end{lstlisting}

The implementation relies on \texttt{numpy}, \texttt{scipy}, and \texttt{scikit-learn}.
Each sampler class implements three main methods inspired from the \texttt{scikit-learn} API:
(i) \texttt{fit} computes the parameter values which are later needed to resample the data into a balanced set;
(ii) \texttt{sample} performs the sampling and returns the data with the desired balancing ratio;
and (iii) \texttt{fit\_sample} is equivalent to calling the method \texttt{fit} followed by the method \texttt{sample}.
A class \texttt{Pipeline} is inherited from \texttt{scikit-learn} toolbox to automatically combine \texttt{samplers}, \texttt{transformers}, and \texttt{estimators}.

\section{Implemented methods}

The \texttt{imbalanced-learn} toolbox provides four different strategies to tackle the problem of imbalanced dataset:
(i) under-sampling, (ii) over-sampling, (iii) a combination of both, and (iv) ensemble learning.
The following subsections give an overview of the techniques implemented.

\subsection{Notation and background}

Let $\chi$ an imbalanced dataset with $\chi_{min}$ and $\chi_{maj}$ being the subset of samples belonging to the minority and majority class, respectively.
The balancing ratio of the dataset $\chi$ is defined as:

\begin{equation}
r_{\chi} = \frac{|\chi_{min}|}{|\chi_{maj}|} \ ,
\end{equation}

\noindent where $|\cdot|$ denotes the cardinality of a set. The balancing process is equivalent to resample $\chi$ into a new dataset $\chi_{res}$ such that $r_{\chi} > r_{\chi_{res}}$.

\subsection{Under-sampling}

Under-sampling refers to the process of reducing the number of samples in $\chi_{maj}$.
The implemented methods can be categorized into 2 groups: (i) fixed under-sampling and (ii) cleaning under-sampling.

\emph{Fixed under-sampling} refer to the methods which perform under-sampling to obtain the appropriate balancing ratio $r_{\chi_{res}}$.
The implemented methods perform the under-sampling based on different criteria such as: (i) random selection, (ii) clustering, (iii) nearest neighbours rule (i.e., \texttt{NearMiss}~\citep{mani2003knn}), and (iv) classification accuracy (i.e., \texttt{instance hardness threshold}~\citep{smith2014instance}).

In the contrary to the previous methods, \emph{cleaning under-sampling} do not allow to reach specifically the balancing ratio $r_{\chi_{res}}$, but rather clean the feature space based on some empirical criteria.
These criteria are derived from the nearest neighbours rule, namely: (i) \texttt{condensed nearest neighbours}~\citep{hart1968condensed}, (ii) \texttt{edited nearest neighbours}~\citep{wilson1972asymptotic}, (iii) \texttt{one-sided selection}~\citep{kubat1997addressing}, (iv) \texttt{neighbourhood cleaning rule}~\citep{laurikkala2001improving}, and (v) \texttt{Tomek links}~\citep{tomek1976two}.

\subsection{Over-sampling}

In the contrary to under-sampling, data balancing can be performed by over-sampling such that new samples are generated in $\chi_{min}$ to reach the balancing ratio $r_{\chi_{res}}$.
Two methods are currently available: (i) \texttt{Random over-sampling} is performed by randomly replicating the samples of $\chi_{min}$ to obtain the appropriate balancing ratio $r_{\chi_{res}}$ and \texttt{SMOTE} which randomly generate new samples between tuple of nearest neighbours of $\chi_{min}$~\citep{chawla2002smote}.
Different variants of this algorithm have been proposed: \texttt{SMOTE borderline 1 \& 2}~\citep{han2005borderline} and \texttt{SMOTE SVM}~\citep{nguyen2011borderline}.


\subsection{Combination of over- and under-sampling}

\texttt{SMOTE} over-sampling can lead to over-fitting which can be avoided by applying cleaning under-sampling methods~\citep{prati2009data}.
In that regard, \cite{batista2003balancing} combined \texttt{SMOTE} either with \texttt{Tomek links} or \texttt{edited nearest neighbours}.

\subsection{Ensemble learning}

Under-sampling methods imply that samples of the majority class are lost during the balancing procedure.
Ensemble methods offer an alternative to use most of the samples.
In fact, an ensemble of balanced sets is created and used to later train any classifier.
Two methods are available to build such ensemble proposed by~\cite{liu2009exploratory}: \texttt{EasyEnsemble} and \texttt{BalanceCascade}.
The former is based on iteratively applying the \texttt{random under-sampling} method to build several sets, each of them with a desired balancing ratio $r_{\chi_{res}}$.
The latter differs such that a classifier is used at each iteration to determine the class of the randomly selected samples.
Misclassified samples are kept and propagated in the next subset.

\section{Future plans and conclusion}

In this paper, we shortly presented the foundations of the \texttt{imbalanced-learn} toolbox vision and API.
As avenues for future works, additional methods based on prototype/instance selection, generation, and reduction will be added as well as additional user guides.


\bibliography{bibtex}

\begin{thebibliography}{21}
\providecommand{\natexlab}[1]{#1}
\providecommand{\url}[1]{\texttt{#1}}
\expandafter\ifx\csname urlstyle\endcsname\relax
  \providecommand{\doi}[1]{doi: #1}\else
  \providecommand{\doi}{doi: \begingroup \urlstyle{rm}\Url}\fi

\bibitem[Batista et~al.(2003)Batista, Bazzan, and Monard]{batista2003balancing}
G.~E. Batista, A.~L. Bazzan, and M.~C. Monard.
\newblock Balancing training data for automated annotation of keywords: a case
  study.
\newblock In \emph{WOB}, pages 10--18, 2003.

\bibitem[Chawla et~al.(2002)Chawla, Bowyer, Hall, and
  Kegelmeyer]{chawla2002smote}
N.~V. Chawla, K.~W. Bowyer, L.~O. Hall, and W.~P. Kegelmeyer.
\newblock {S}{M}{O}{T}{E}: synthetic minority over-sampling technique.
\newblock \emph{Journal of artificial intelligence research}, pages 321--357,
  2002.

\bibitem[Dal~Pozzolo et~al.(2013)Dal~Pozzolo, Caelen, Waterschoot, and
  Bontempi]{dal2013racing}
A.~Dal~Pozzolo, O.~Caelen, S.~Waterschoot, and G.~Bontempi.
\newblock Racing for unbalanced methods selection.
\newblock In \emph{International Conference on Intelligent Data Engineering and
  Automated Learning}, pages 24--31. Springer, 2013.

\bibitem[Han et~al.(2005)Han, Wang, and Mao]{han2005borderline}
H.~Han, W.-Y. Wang, and B.-H. Mao.
\newblock Borderline-smote: a new over-sampling method in imbalanced data sets
  learning.
\newblock In \emph{International Conference on Intelligent Computing}, pages
  878--887. Springer, 2005.

\bibitem[Hart(1968)]{hart1968condensed}
P.~Hart.
\newblock The condensed nearest neighbor rule.
\newblock \emph{Information Theory, IEEE Transactions on}, 14\penalty0
  (3):\penalty0 515--516, May 1968.

\bibitem[He and Garcia(2009)]{he2009learning}
H.~He and E.~Garcia.
\newblock Learning from imbalanced data.
\newblock \emph{Knowledge and Data Engineering, IEEE Transactions on},
  21\penalty0 (9):\penalty0 1263--1284, 2009.

\bibitem[Kubat et~al.(1997)Kubat, Matwin, et~al.]{kubat1997addressing}
M.~Kubat, S.~Matwin, et~al.
\newblock Addressing the curse of imbalanced training sets: one-sided
  selection.
\newblock In \emph{International Conference in Machine Learning}, volume~97,
  pages 179--186. Nashville, USA, 1997.

\bibitem[Kuhn(2015)]{kuhn2015caret}
M.~Kuhn.
\newblock Caret: classification and regression training.
\newblock \emph{Astrophysics Source Code Library}, 1:\penalty0 05003, 2015.

\bibitem[Laurikkala(2001)]{laurikkala2001improving}
J.~Laurikkala.
\newblock \emph{Improving identification of difficult small classes by
  balancing class distribution}.
\newblock Springer, 2001.

\bibitem[Liu et~al.(2009)Liu, Wu, and Zhou]{liu2009exploratory}
X.-Y. Liu, J.~Wu, and Z.-H. Zhou.
\newblock Exploratory undersampling for class-imbalance learning.
\newblock \emph{IEEE Transactions on Systems, Man, and Cybernetics, Part B
  (Cybernetics)}, 39\penalty0 (2):\penalty0 539--550, 2009.

\bibitem[Mani and Zhang(2003)]{mani2003knn}
I.~Mani and I.~Zhang.
\newblock knn approach to unbalanced data distributions: a case study involving
  information extraction.
\newblock In \emph{Proceedings of Workshop on Learning from Imbalanced
  Datasets}, 2003.

\bibitem[Nguyen et~al.(2011)Nguyen, Cooper, and Kamei]{nguyen2011borderline}
H.~M. Nguyen, E.~W. Cooper, and K.~Kamei.
\newblock Borderline over-sampling for imbalanced data classification.
\newblock \emph{International Journal of Knowledge Engineering and Soft Data
  Paradigms}, 3\penalty0 (1):\penalty0 4--21, 2011.

\bibitem[Pedregosa et~al.(2011)Pedregosa, Varoquaux, Gramfort, Michel, Thirion,
  Grisel, Blondel, Prettenhofer, Weiss, Dubourg, et~al.]{pedregosa2011scikit}
F.~Pedregosa, G.~Varoquaux, A.~Gramfort, V.~Michel, B.~Thirion, O.~Grisel,
  M.~Blondel, P.~Prettenhofer, R.~Weiss, V.~Dubourg, et~al.
\newblock Scikit-learn: {M}achine learning in python.
\newblock \emph{Journal of Machine Learning Research}, 12\penalty0
  (Oct):\penalty0 2825--2830, 2011.

\bibitem[Prati et~al.(2009)Prati, Batista, and Monard]{prati2009data}
R.~C. Prati, G.~E. Batista, and M.~C. Monard.
\newblock Data mining with imbalanced class distributions: concepts and
  methods.
\newblock In \emph{Indian International Conference Artificial Intelligence},
  pages 359--376, 2009.

\bibitem[Rastgoo et~al.(2016)Rastgoo, Lemaitre, Massich, Morel, Marzani,
  Garcia, and Meriaudeau]{rastgoo2016tackling}
M.~Rastgoo, G.~Lemaitre, J.~Massich, O.~Morel, F.~Marzani, R.~Garcia, and
  F.~Meriaudeau.
\newblock Tackling the problem of data imbalancing for melanoma classification.
\newblock In \emph{Bioimaging}, 2016.

\bibitem[Smith et~al.(2014)Smith, Martinez, and
  Giraud-Carrier]{smith2014instance}
M.~R. Smith, T.~Martinez, and C.~Giraud-Carrier.
\newblock An instance level analysis of data complexity.
\newblock \emph{Machine learning}, 95\penalty0 (2):\penalty0 225--256, 2014.

\bibitem[Sonnenburg et~al.(2010)Sonnenburg, Henschel, Widmer, Behr, Zien, Bona,
  Binder, Gehl, Franc, et~al.]{sonnenburg2010shogun}
S.~C. Sonnenburg, S.~Henschel, C.~Widmer, J.~Behr, A.~Zien, F.~de Bona,
  A.~Binder, C.~Gehl, V.~Franc, et~al.
\newblock The {S}{H}{O}{G}{U}{N} machine learning toolbox.
\newblock \emph{Journal of Machine Learning Research}, 11\penalty0
  (Jun):\penalty0 1799--1802, 2010.

\bibitem[Tomek(1976)]{tomek1976two}
I.~Tomek.
\newblock Two modifications of {C}{N}{N}.
\newblock \emph{Systems, Man, and Cybernetics, IEEE Transactions on},
  6:\penalty0 769--772, 1976.

\bibitem[Torgo(2010)]{torgo2010data}
L.~Torgo.
\newblock \emph{Data mining with {R}: learning with case studies}.
\newblock Chapman \& Hall/CRC, 2010.

\bibitem[Wilson(1972)]{wilson1972asymptotic}
D.~L. Wilson.
\newblock Asymptotic properties of nearest neighbor rules using edited data.
\newblock \emph{Systems, Man and Cybernetics, IEEE Transactions on}, \penalty0
  (3):\penalty0 408--421, 1972.

\bibitem[Yang and Wu(2006)]{yang200610}
Q.~Yang and X.~Wu.
\newblock 10 challenging problems in data mining research.
\newblock \emph{International Journal of Information Technology \& Decision
  Making}, 5\penalty0 (04):\penalty0 597--604, 2006.

\end{thebibliography}

\end{document}